\documentclass[letterpaper, 10pt, conference]{ieeeconf}
\pdfoutput=1

\IEEEoverridecommandlockouts                          
\usepackage{multirow}
\usepackage{float}
\usepackage{graphicx}
\graphicspath{ {./images/} }
\usepackage{xcolor}
\usepackage [english]{babel}
\usepackage [autostyle, english = american]{csquotes}
\MakeOuterQuote{"} 
\usepackage{amsthm}
\newtheorem*{problem}{Problem}
\newtheorem*{rem}{Remark}

\theoremstyle{definition}
\newtheorem*{exmp}{Example}

\usepackage{amsmath, amssymb}

\DeclareMathOperator*{\argmax}{arg\,max}

\title{\LARGE \bf
Context-Dependent Anomaly Detection with Knowledge Graph Embedding Models
}

\author{Nathan Vaska$^{1}$, Kevin Leahy$^{1}$, and Victoria Helus$^{1}$% <-this % stops a space
\thanks{DISTRIBUTION STATEMENT A. Approved for public release. Distribution is unlimited. This material is based upon work supported by the Under Secretary of Defense for Research and Engineering under Air Force Contract No. FA8702-15-D-0001. Any opinions, findings, conclusions or recommendations expressed in this material are those of the author(s) and do not necessarily reflect the views of the Under Secretary of Defense for Research and Engineering. © 2022 Massachusetts Institute of Technology. Delivered to the U.S. Government with Unlimited Rights, as defined in DFARS Part 252.227-7013 or 7014 (Feb 2014). Notwithstanding any copyright notice, U.S. Government rights in this work are defined by DFARS 252.227-7013 or DFARS 252.227-7014 as detailed above. Use of this work other than as specifically authorized by the U.S. Government may violate any copyrights that exist in this work.
}% <-this % stops a space
\thanks{$^{1}$Authors are with MIT Lincoln Laboratory, Lexington, MA 02420}
\thanks{Corresponding author: {\tt\small victoria.helus@ll.mit.edu}}%
}

\begin{document}

\maketitle
\thispagestyle{empty}
\pagestyle{empty}

%%%%%%%%%%%%%%%%%%%%%%%%%%%%%%%%%%%%%%%%%%%%%%%%%%%%%%%%%%%%%%%%%%%%%%%%%%%%%%%%
\begin{abstract}

Increasing the semantic understanding and contextual awareness of machine learning models is important for improving robustness and reducing susceptibility to data shifts. In this work, we leverage contextual awareness for the anomaly detection problem. Although graphed-based anomaly detection has been widely studied, context-dependent anomaly detection is an open problem and without much current research. We develop a general framework for converting a context-dependent anomaly detection problem to a link prediction problem, allowing well-established techniques from this domain to be applied. We implement a system based on our framework that utilizes knowledge graph embedding models and demonstrates the ability to detect outliers using context provided by a semantic knowledge base. We show that our method can detect context-dependent anomalies with a high degree of accuracy and show that current object detectors can detect enough classes to provide the needed context for good performance within our example domain.

\end{abstract}

%%%%%%%%%%%%%%%%%%%%%%%%%%%%%%%%%%%%%%%%%%%%%%%%%%%%%%%%%%%%%%%%%%%%%%%%%%%%%%%%
\section{Introduction}

Machine learning approaches today have achieved impressive and at times superhuman performance at a variety of tasks, such as game playing, pattern recognition, classification, and more \cite{He2015DelvingDI, Silver2016MasteringTG, Fuchs2021SuperHumanPI, Dodge2017ASA}; however, such performance is often limited to a narrow domain, with system capability degrading rapidly as data distributions shift away from the training distribution \cite{koh2021wilds, 10.1016/j.patcog.2011.06.019} or inputs are perturbed and corrupted \cite{szegedy2013intriguing, hendrycks2019benchmarking}. As machine learning systems proliferate, gain greater influence over decision making, and increasingly integrate into daily life it becomes imperative to broaden model capabilities.

In contrast, humans maintain an awareness of context that reduces susceptibility to these problems; we reason about the relationships between objects, determine if something belongs, and adjust our beliefs given what we perceive. Aspiring to systems that are expected to explore and navigate through unseen environments or operate in shifting domains demand a higher level understanding; this motivates the exploration of techniques that incorporate contextual relationships as a prior.    

Consider an embodied agent placed in a house and given a high-level instruction such as "I am watching TV, please bring me a spoon \cite{mattersim} for my food, and tell me if you see anything strange." While this may sound simple to a human, a high degree of implicit knowledge is required to ensure successful comprehension and execution of such an instruction. An agent would have to understand the instruction (language processing), flag unknown objects (uncertainty quantification), recognize where it is (scene classification), understand what doesn't belong in its current location (anomaly detection), figure out where a spoon would be and how to get there, where you would be watching TV and how to get there (reasoning), and physically reach the necessary locations (navigation). 

\begin{figure*}[t]
\centering
    \includegraphics[width=2.0\columnwidth]{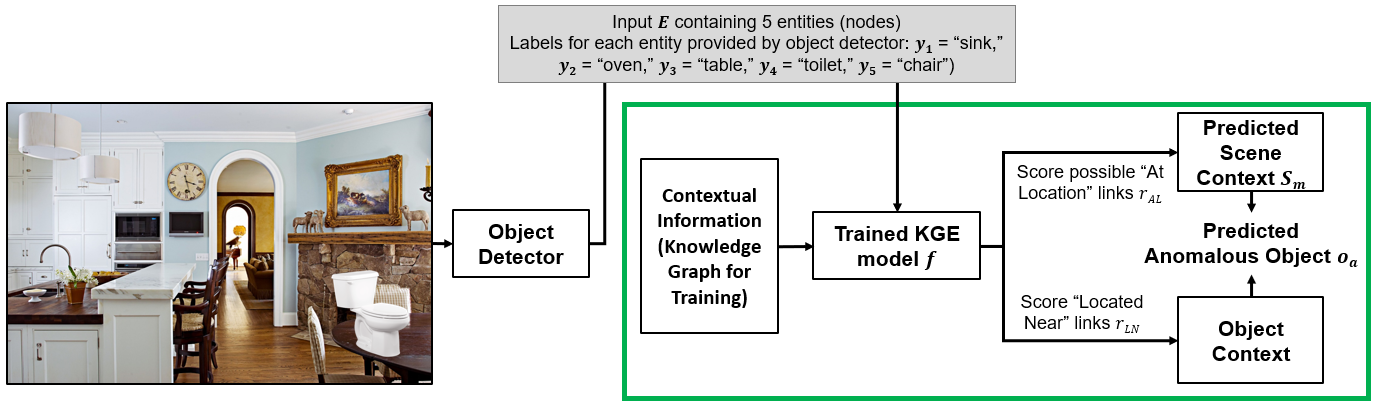}
\begin{minipage}[b]{0.7\textwidth}
    \caption{A notional deployed system. The object detector feeds labels into an anomaly detection system denoted in green, which is the focus of this paper. Notation shown is explained in Section \ref{prob}.}
    \label{fig:conops}
\end{minipage}
\vspace{-0.6cm}
\end{figure*}

We choose to focus initially on the anomaly detection component of this problem. Detecting anomalies within a scene requires a more generalized contextual understanding than current narrow approaches can offer; a conceptual system would operate as shown in Fig. \ref{fig:conops}. For example, while a teapot may not be out of place in a kitchen or a living room, it would be considered odd in a bathroom, even if the teapot itself is not strange. 

Given the lack of much prior work on \textit{context-dependent} anomaly detection \cite{Pang2021DeepLF, Chandola2009AD}, the design space of models addressing this problem is largely unexplored. In this work we treat the context-dependent anomaly detection problem as an extension of the link prediction task (also known as graph completion) \cite{Akoglu2014GraphBA, Rossi2020KnowledgeGE}, a popular problem that has multiple benchmarks \cite{Bordes2013TranslatingEF, Toutanova2015OV} and applications in disease monitoring, recommendation systems, drug discovery, social media suggestions, and more. Our contributions in this work are the following:
\begin{itemize}
    \item We develop a general framework for mapping any context-dependent anomaly detection to the link prediction task for which a knowledge graph exists or can be constructed.
    \item We apply our method to a specific domain, identifying anomalous objects within household scenes, and show that our method achieves good performance while remaining efficient and interpretable.
    \item We show that our method could be implemented on an embodied agent equipped with widely-available object detectors.
\end{itemize}

\section{Related Work}

\subsection{Anomaly Detection}
Anomaly detection is the task of detecting outliers from a given data distribution, and is relevant in many fields and applications, including cybersecurity, finance, healthcare, surveillance, and more. Deep learning methods have demonstrated promising performance in detecting anomalies \cite{Pang2021DeepLF}; yet, most of the existing techniques are focused on \textit{point anomalies} (an individual instance that lies far from the rest of the instances), rather than the more challenging \textit{group anomalies} (an individual instance may be considered normal, but a collection of instances is anomalous compared to the other instances), or \textit{contextual anomalies} (an instance can be normal or abnormal depending on its current context) \cite{Pang2021DeepLF, Chandola2009AD}. Performing context-dependent anomaly detection usually involves trying to reduce the problem to a point anomaly detection problem, where multiple models may be conditioned on specific contexts, or training a model on previous data (often sequential or time-series) to predict expected behavior that can then be flagged as anomalous or not \cite{Chandola2009AD, Bozcan2021ContextDependentAD}. In contrast, our method takes advantage of the full context and does not require historical patterns to train on or compare to. 

Another interesting and relevant early approach demonstrates the ability to detect abnormal nodes in applications that can be modeled as bipartite graphs \cite{Sun2005NF}; currently, detecting anomalies in graphs (anomalous nodes, edges, or sub-graphs) is a widely researched problem \cite{Akoglu2014GraphBA, Ma2021ACS}, often used in applications such as social networks and transportation networks. While these approaches are looking for anomalies in already completed graphs, our focus on extending the link prediction task assumes an incomplete graph. 

\subsection{Visual Question Answering}
Visual question answering (VQA) is the task of being able to provide an answer to a natural language question based on image data \cite{Agrawal2015VQAVQ}. Much like the embodied problem posed in the introduction, even answering questions with a simple "yes" or "no" requires a nontrivial implicit thought process. Unlike other related language and image tasks (e.g., image captioning), VQA often demands a specificity in answers that can only be accomplished through the type of reasoning that requires human-aligned knowledge and understanding. Unlike the problem we are considering, VQA solvers are usually focused on answering more specific questions that often even call out features in an image (e.g., "how many balls are to the left of the blue cube?"), rather than answering more general questions such as "is there anything unusual in this scene?" Several VQA datasets have been developed \cite{Agrawal2015VQAVQ, Yi2020CLEVRERCE, Johnson2017CLEVRAD, Krishna2016VisualGC, Lin2014MicrosoftCC} and include sentence descriptions of the images or annotated ground-truth of the scene and objects. For our work, we leverage one of these datasets for its annotations and relationship-rich scenes to train instead for context-dependent anomaly detection. 

\subsection{Vision-and-Language Navigation}
Similar to VQA, vision-and-language navigation (VLN) seeks to connect language processing capabilities to navigate to a goal. This is also currently a very active field of research \cite{Wu2021VisualandLanguageNA}, although most current systems require highly detailed instructions (e.g., "go down the hallway, turn at your second left, enter the first door to the right where there is a bed in the room.") rather than high-level instructions (e.g., "go to the bedroom"). In addition, generalizing to unseen environments is also an open problem \cite{mattersim}. Extending this to an environment that even a human has not seen and asking for an agent to use context to follow high-level instructions is an even more difficult task. Our work connects to this task through its leveraging of common-sense knowledge, a feature which will likely be necessary for generalization to unseen environments.

\section{Problem Formulation}\label{prob}
To utilize a standard link prediction model for context-dependent anomaly detection on an input, the input must be framed as a graph. The following  section provides general framework for this conversion, after which we will provide an example of this process applied to a specific domain.

\subsection{Entities and Relationships}

We introduce the idea of an \emph{entity}, which represents any element or attribute in an environment of interest. We frame the anomaly detection problem as one of correctly predicting relationships between entities. An anomaly occurs when an entity appears among a set of entities with which it doesn't have a relationship in common. To formalize this notion of anomalous entities, %For the purposes of our formulation,
we attach several properties to each of our entities to help characterize our problem:

Let $T$ be the set of \textit{types} of entities. Types represent the broad classes of entities that we wish to reason about. Individual entities of the same type are distinguishable by labels. Let $Y$ be the set of \textit{true labels} that the entities are representing. Subsets $Y_{t_i} \subseteq Y$ are divided by entity type such that $\bigcup_{t \in T} Y_t = Y$. Let $R$ be the set of \textit{types of relationships} between two entity types (self-relation included).

Then let $E$ be the set of entities that represent an input such that $e_i \in E, e_i := (y_i, t_i, \{R_{t_i,t_j}\})$, where $y_i \in Y_{t_i}$ is the known true label of the entity, $t_i$ is the known type of entity and $R_{t_i,t_j}$ is the known set of possible relationships between entity types $t_i, t_j \in T$. This set provides the context of the input.

\begin{exmp} To ground this formulation in a more tangible problem, we apply it to detecting anomalous objects within household scenes.

Since we wish to reason about objects in scenes, let 
$T = \{t_o, t_s\}$ be the set of types of possible entities---object or scene.
Then $Y$ is the full set of possible labels across all entity types (any object or scene label),
with $Y_{t_s} \subset Y$ being the set of possible scene ($t_s$) labels in our domain (e.g., "office," "bathroom," "kitchen"). Similarly, $Y_{t_o}\subset Y$ is the set of possible object ($t_o$) labels in our domain (e.g., "toaster," "desk," "oven").

In this example, objects can belong in a scene or be associated with other objects. Let $R = \{r_{AL}, r_{LN}\}$ be the set of the possible relationships "At Location" and "Located Near" that each link $l$ can represent. Then $R_{t_o,t_s} = \{r_{AL}\}$ and $R_{t_o,t_o} = \{r_{LN}\}$. We do not include any scene-to-scene relations so $R_{t_s,t_s} = \{\emptyset\}$. One might expect the "toaster" and "oven" objects to share the "Located Near" relationship, and each of them to share the "At Location" relationship with "kitchen." On the other hand, "desk" is unlikely to share any of these relationships.

\end{exmp}

\subsection{Knowledge Graphs and Link Prediction}

A knowledge graph, as defined in \cite{Hogan2021KnowledgeG}, is "a graph of data intended to accumulate and convey knowledge of the real world, whose nodes represent entities of interest and whose edges represent relations between these entities." Knowledge graphs allow information from multiple data sources to be flexibly integrated into a cohesive schema, making it a powerful tool for describing context and contextual anomalies. Furthermore, their interpretability offers an advantage when there is a need to understand a system's outputs. 

We define a directed graph $G = (E, L)$ as a set of entities $E$ (also known as vertices or nodes) that are connected by a set of relationships (i.e., links) $L \subseteq E\times E$, where each $l \in L$ connects what is referred to as a "head" entity and a "tail" entity. Knowledge graph entities can also contain related characteristics referred to as attributes. Attributes can vary depending on entity type (e.g., an object entity might have an attribute describing weight, but an audio entity would not have such an attribute), and can be modeled as a relationship or an entity. We do not include attributes in our example, but the framework can easily accommodate them. 

The link prediction problem takes an incomplete graph (a graph with at least one pair of vertices that do not have a connecting edge) as its training data. It assumes that the observed links are a subset of the true links that should exist. The goal is to try to predict the existence of the nonexistent true link(s). A link prediction model is a model $f: E \times E \rightarrow \mathbb{R}$ that is trained to map potential links to a score, where a high score is indicative of a true link and a low score a corrupted link (i.e., a link that should not exist). Corrupted links are generated by permuting a true link's head or tail entity such that the resulting head, relation, tail triplet does not exist in the training data. For a complete description of training a link prediction model, see \cite{Bordes2013TranslatingEF}.

Given a trained link prediction model $f$, then for a new link $l=(e_i,e_j) \notin L$ we can calculate a score $f(e_i, e_j)$, where $i \neq j$. If $f(e_i, e_j)$ is high then link $l$ is likely to exist while if $f(e_i, e_j)$ is low $l$ is unlikely to exist.

\begin{problem}
Given a set of $n$ entities $[e_1, e_2, ..., e_n] \in E$, of which one entity is known to be have a true weak or nonexistent relationship with the other entities, identify this entity as anomalous.
\end{problem}
\section{Approach}
We use the link prediction model to score possible links to perform context-dependent anomaly detection on an input. e can consider a particular object's strength of relationship with the set of other objects $O$ and the set of possible scenes $S$ as follows: given $o_i \in O, \forall o_j \in O, i \neq j$ or $\forall s_j \in S_m$ and $\forall r \in \{R_{t_i, t_j}\}$, a set of scores can be calculated for each relationship type $Z_r = \{f_r(o_i, o_j)\}$ (or $\{f_r(o_i, s_j)\}$ for object-scene relations). As these are multiple separate scores and the goal is to have one final anomaly score to assess a candidate object, the sets of relationship scores can be aggregated to give an overall of anomaly score $Z = g(h_{r_1}(Z_{r_1}), h_{r_2}(Z_{r_2}), ..., h_{r_n}(Z_{r_n}))$, where $g$ is a domain-specific aggregation function and each $h_r \forall r \in \{R_{t_i, t_j}\}$ is a relationship specific aggregation function. Higher values of $Z$ indicate that entity $o_i$ is more likely to be anomalous based on the context given by $O$ and $S$. Note that individual relationship types may have either a positive or negative weight in $g$; a negative weight for a relationship type implies that entities with weak links of that type are more likely to be anomalous and vice versa. This allows any type of relationship to be included in the graph, as long as the sign of its contribution to the anomaly score is constant across all entities. Additionally, note that different types of anomalies can be detected by pairing the same graph and link prediction model with multiple aggregation functions $g$.

Given a model $f$ trained to predict "At Location" and "Located Near" links ($r_{AL}$ and $r_{LN}$, respectively), the following procedure can be used to compute object and scene context scores and perform anomaly inference.

\subsection{Object Context}

Let $O$ be the set of entities that are objects in a given input (i.e., $o_i = (y_i, t_o, \{R_{t_o, t_o}, R_{t_o, t_s}\}) \in O$), where the input's object labels $y_i$ are known.

For a particular set $O$ of room objects and injected anomalous object, given an anomaly candidate $o_{ca}\in O$ from the set, we can calculate a "Located Near" score $f_{r_{LN}}(o_{ca}, o_i)$ between the anomaly candidate and every other object in the set. The average of these scores, $z_o$, is the "object score." It is a measure of how likely the anomaly candidate is to be found in the same scene as the other objects in the set and can be calculated by:

\begin{equation}
    z_o(o_{ca}) = \frac{1}{|O|}\sum_{e \in O \backslash{o_{ca}}} f_{r_{LN}}(o_{ca}, o)
    \label{eq:first}
\end{equation}

\subsection{Scene Context}
 Next, we are interested in predicting the input's scene based off of the set of objects, so that the predicted scene can be used as additional contextual information to help identify the anomaly. Let $S$ be the set of all \textit{possible} scene type entities such that $s \in S = (y, t_s, \{R_{t_o, t_s}\}) \forall y$ in $Y_{t_s}$ and let $S_m$ be the set of entities that represent the $m$ most likely scenes for a given input (i.e., $s_i = (y_i, t_s, \{R_{t_o, t_s}\}) \in S_m$) and $|S_m| = m$.
 
 We can calculate a "scene score" via the following method. First, generate an "At Location" score $f_{r_{AL}}(o_i, s_j)$ between each object (including the anomaly candidate) and each scene from the set of scene candidates $S$. Then, average the scores across objects to provide a compatibility score between each scene type and the set of objects. Select the top $m$ scenes based on compatibility score as likely candidates and call this set $S_m$, which can be written as:
 
 \begin{equation}
    S_m = \argmax_{S' \subseteq S, |S'|= m, s \in S} \frac{1}{|O|}*\sum_{o \in O} f_{r_{AL}}(o, s)
\end{equation}

Similar to above, the "At Location" scores from the scene types in $S_m$ are then averaged to get the scene score $z_s$, which indicates how likely the current anomaly candidate object is to be found in the most likely scene types for the given set $O$.
 
 \begin{equation}
    z_s(o_{ca}) = \frac{1}{m}\sum_{s \in S_m} f_{r_{AL}}(o_{ca}, s)
\end{equation}
 
 Selecting the wrong scenes for $S_m$ is always a risk, and possibly a highly impactful one\textemdash something like a chainsaw may be innocuous in a basement location, but alarming in an office. Selecting more than one scene for $S_m$ helps mitigate the chance that incorrect scene predictions will solely be used to generate the scene context. Since there are often several scene types with similar traits (i.e., garden, yard) using multiple scenes as context is not expected to overly dilute information provided by correct scene type predictions. 

\subsection{Aggregating Context and Computing Anomaly Score}
The overall anomaly score $z$ for the current anomaly candidate is calculated by taking the negative weighted sum of $z_o$ and $z_s$ where $\alpha \in [0, 1]$ is a weighting parameter. Note that as "At Location" and "Located Near" scores are always generated by the same model and averaged over their variable inputs, they do not require any additional normalization.

\begin{equation}
    z(o_{ca})= -\alpha*z_o(o_{ca}) - (1 - \alpha)*z_s(o_{ca})
    \label{eq:last}
\end{equation}

A anomaly score can be calculated for each object in the the set as an anomaly candidate; the anomalous object $o_a$ can then be predicted by simply taking an $\argmax$ over all anomaly scores as per Eq. \ref{eq:actuallast}.

\begin{equation}
    o_a = \argmax_{o \in O} z(o)
    \label{eq:actuallast}
\end{equation}

An advantage of this procedure is its computational efficiency when sufficient memory is available. Calculating the link scores for every relation type is the most computationally heavy component, taking $O(L|R||E|^2)$ time where $L$ represents the time complexity of of the link scoring method for the chosen link prediction model. However, once calculated, the scores can be reused indefinitely for any number of inputs. Assuming that inference method consists of simple sum and average operations on the score table, as is the case for our example domain, the time complexity of performing anomaly inference on a single input is $O(|R||E_i|^2)$ and on many inputs is $O(N|R||E_i|^2)$ where $N$ is the number of inputs and $E_i$ is the largest set of entities considered in any input. Assuming that $|R| \ll N$ and $|E_i| \ll N$, the runtime on many inputs reduces to $O(N)$. In practice we found that, using a standard CPU, we were able to complete inference on our largest anomaly dataset on the order of minutes. The results of our process are detailed in Section \ref{results}.

\begin{rem}
While we did not explore this feature in depth in this work, because an explicit anomaly score is calculated between each object and scene, our model allows the factors which led to a particular object being classified as anomalous to be examined more closely than is possible with a standard neural network classification architecture.
\end{rem}

\section{Experimental Methods}

\subsection{Model}
We used a family of knowledge graph embedding (KGE) models for our link prediction models in our experiments \cite{Rossi2020KnowledgeGE, Wang2021ASKG, Nguyen2017AnOO}. KGE models are the first choice for graph completion; since graphs can contain millions of links, model efficiency is extremely important. KGE models address this problem by learning an N-dimensional vector to represent each graph entity. Link prediction is treated as a comparison between entity vectors, often using another set of relationship vectors to differentiate between different graph relations. As this comparison is highly computationally efficient, KGE models can process even large knowledge graphs within a reasonable period of time. Additionally, since the training of non-neural KGE models often produces a descriptive set of entity embeddings it also provides more insight into the learning process than other methods. Popular models are also easily accessible through the TorchKGE package \cite{Boschin2020TorchKGEKG}, which we used in our experiments. 

Additionally, since the (non-neural) KGE model training process often produces a descriptive set of entity embeddings it also provides more insight into the learning process than other methods. Popular models are also easily accessible through the TorchKGE package \cite{Boschin2020TorchKGEKG}, which we used in our experiments. 

\subsection{Data}

Although there are general knowledge graphs available, to the best of our knowledge there are none that contain a large amount of realistic object and scene information with both "At Location" and "Located Near" relations. Furthermore, there are no publicly available datasets containing realistic anomalies for common household scenes. The following sections detail our methodology for creating these datasets.

\subsubsection{Object-Scene Dataset Methodology}

\begin{figure}[t]
\centering
\begin{minipage}[b]{.47\textwidth}
     \centering
    \includegraphics[width=1.0\columnwidth]{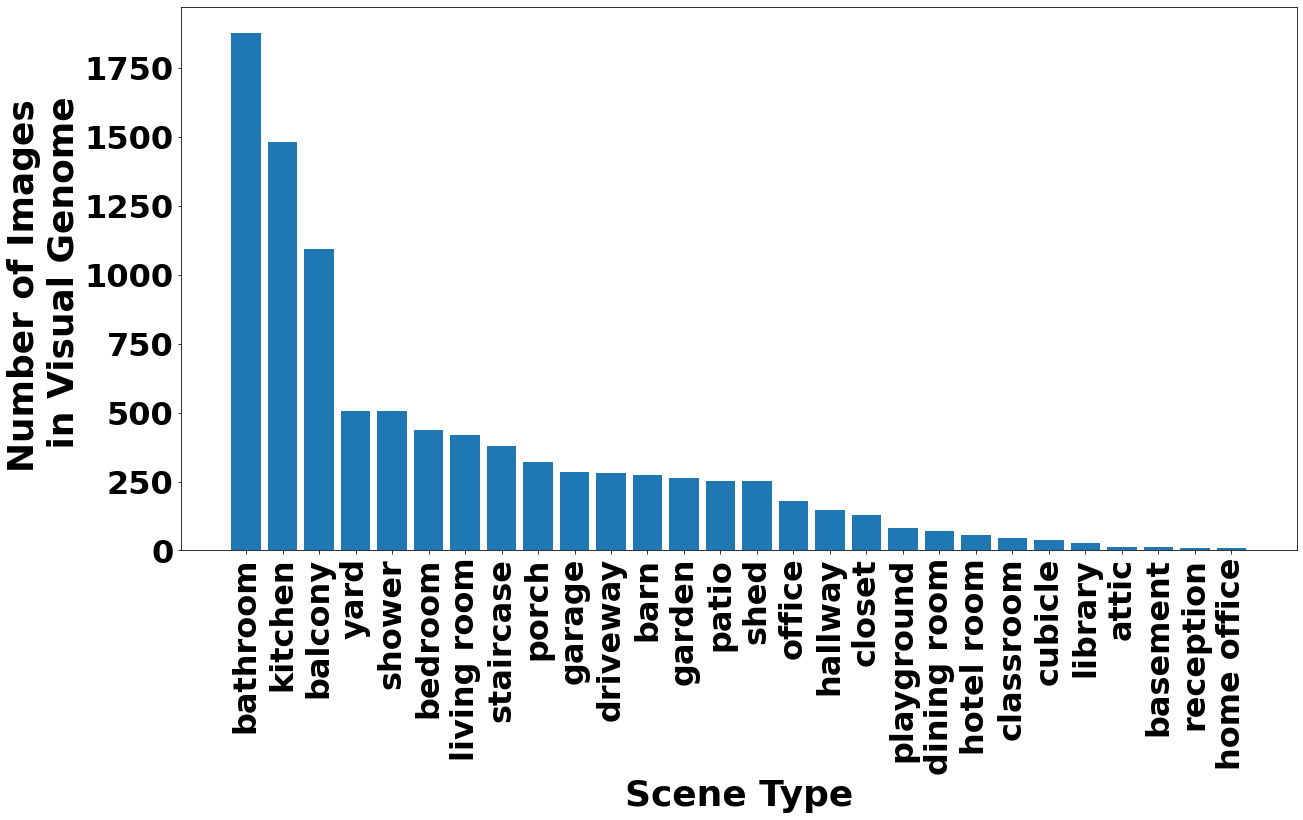}
    \caption{Number of Visual Genome Images Per Scene Type. Note that scenes from bathroom, kitchen, and balcony far outnumber scenes from other categories.}
    \label{fig:scenedists}
\end{minipage}\hfill
\end{figure}
\begin{figure}
\begin{minipage}[b]{.47\textwidth}
    \includegraphics[width=1.0\columnwidth]{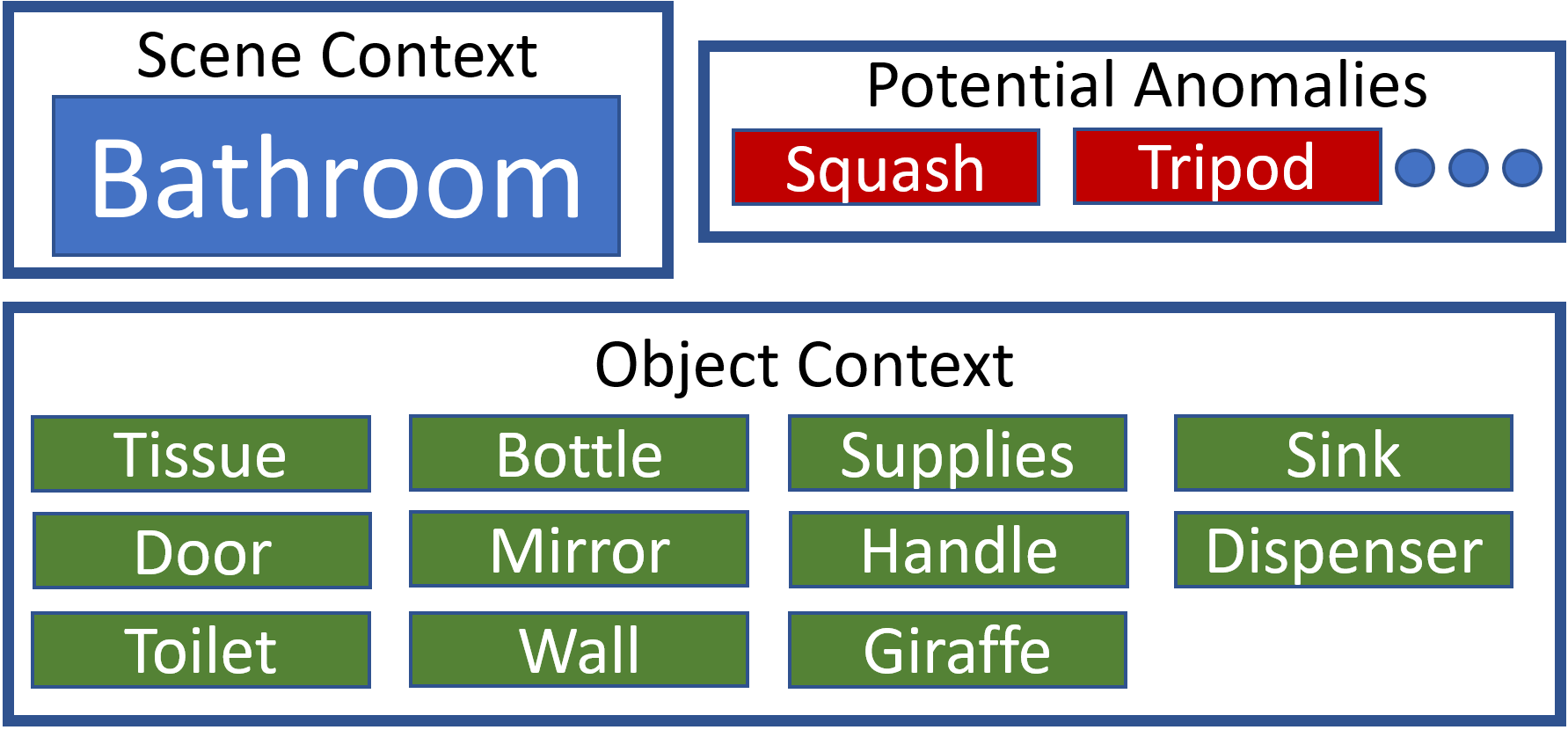}
    \caption{Example drawn from the Detector version of the Unique Out-of-Scene anomaly dataset. Blue represents the scene type, green the labeled objects, and red the possible anomalies for this scene type, of which two examples are shown. To form an individual anomaly datapoint, one anomaly is selected (i.e., squash) and links are hypothesized to exist between the scene, each object, and the anomaly.}
    \label{fig:anomaly-example}
\end{minipage}
\vspace{-0.5cm}
\end{figure}

\begin{table}[t]
\parbox{1.0\linewidth}{
        \centering
        \begin{tabular}{ |l|c|c|c|c|c|  }
            %  \hline
            %  \multicolumn{4}{|c|}{Country List} \\
             \hline
             KG Dataset & Entities & At Location & Located Near & Total \\
             \hline
             Full      & 4,911 & 21,972 & 2,787,736 & 2,809,708 \\
             Filtered  & 1,535 & 15,747 & 65,048    & 80,795 \\
             Detector  &   504 & 4,883  & 119,432   & 124,315\\
             \hline
            \end{tabular}
            \centering 
            \caption{Training Knowledge Graph Link Counts and Distribution between link types for each dataset}
            \vspace{-0.8cm}
            \label{table:tdataset}
        \hfill
        }
\end{table}
\begin{table}
\parbox{1.0\linewidth}{
        \centering
        \begin{tabular}{|l|c|c|c| }
             \cline{3-4}
             \multicolumn{2}{c|}{} & \multicolumn{2}{|c|}{Number of Datapoints}   \\
             \hline
             Anomaly Dataset & Average Objects & Out & Unique Out \\
             \hline
             Full & 20.55 & 934,202 & 577,907 \\
             \hline
             Filtered & 19.31 & 374,187 & 136,937  \\
             \hline
             Detector & 9.31 & 96,817 & 47,002  \\
             \hline
            \end{tabular}
            \centering
            \caption{Anomaly dataset information for each dataset and anomaly filter }
            \vspace{-1.0cm}
            \label{table:a-dataset}
        }
\end{table}

 A subset of the Visual Genome \cite{Krishna2016VisualGC} dataset was used as the main source of data for the object-scene knowledge graph. Visual Genome is a heavily annotated image dataset that is used to train AI systems to have greater understanding of the relationship between objects, regions, and human descriptions in images. It contains about 108,000 images of real-world scenes, of which 8,419 relate to household scenes within twenty-eight, unbalanced categories. The full set of scenes considered and their distribution is shown in Fig. \ref{fig:scenedists}. Critically, Visual Genome images are labeled with both their scene type and the set of objects found in the scene. These labels allowed us to quickly extract semantic information about the relationships between objects, other objects, and scene types from the images; if two objects were found in the same image, a "Located Near" link between the two objects was added to the knowledge graph and if an object was found in a picture labeled as a specific scene type, an "At Location" link was added. To augment the graph, we also incorporated "At Location" and "Located Near" links from ConceptNet, a large and publicly available knowledge graph with 34 million assertions over a variety of link types \cite{Speer2017ConceptNet5A}. While ConceptNet only contributed an additional 151 "At Location" links, a small fraction of our final knowledge graph, ConceptNet also served as an object filter on the raw Visual Genome object sets, as Visual Genome's human annotations are largely unsupervised and include misspellings, pluralization, and other nonsensical objects. By only including objects that exist in both ConceptNet and Visual Genome, we were able to remove the majority of these spurious objects.
 
 The final graph consists of nodes representing both objects and scenes, "At Location" links connecting objects to scenes where they are commonly found, and "Located Near" links connecting two objects that are often found in the same scene. In addition to the "full" dataset we created two filtered versions of the knowledge graph. The first graph filtered object entities by relative scene co-occurence and relative object co-occurrence frequency to eliminate object-scene and object-object connections that appear only infrequently in the data. This subset was intended to provide a less noisy dataset compared to the full, unfiltered dataset at the cost of fewer training examples, and will be referred to as the "Filtered" dataset. The second filtered knowledge graph was created by removing all entities and associated links that did not have a corresponding class in the Google Open Images \cite{OpenImages2} object detection dataset; this subset was intended to more closely match widely-available object detector performance in terms of unique entities that are available for detection. We will refer to this dataset as the "Detector" dataset. Details on all datasets can be found in Table \ref{table:tdataset}.

\subsubsection{Anomaly Dataset Methodology}

There are no publicly available resources detailing anomalous objects for scene types, so we again utilized Visual Genome images to build our anomaly dataset. We defined an anomalous object for a given scene type to be an object that did not appear in a single image of that scene type within Visual Genome's household scene images. We then identified the set of objects found within each scene type, and removed them from the full set of objects to create each scene's anomalous object set. For each annotated image of a specific scene type within Visual Genome, anomalous scenes were created by appending each object from the corresponding anomalous objects list for that scene to the list of labeled objects in that annotated image. We will refer to the full anomaly dataset created through this process as the "Out-of-Scene" anomaly dataset, or "Out" for short. However, given the limited number of images available for some scene types, we were concerned that some non-anomalous objects would be incorrectly added as an anomaly in the testing set. To address this concern, we filtered out any objects that occurred in multiple rooms from the set of anomalous objects. As an example, let's assume we do not have very many images of kitchens in our training set, and none of the images we do have contain chicken meat, which is non-anomalous in a kitchen. However, our training set does contain a lot of images of dining rooms and backyards, and chicken meat appears in images from both these scene types (e.g., at the table in the dining room, on a grill in the backyard); we then choose to omit the "chicken" object from the set of anomalous objects used to create the testing set. Since objects that are only found in one room are more likely to be strongly associated with that room, this process of removing possible spurious anomalies made this "Unique Out-of-Scene" ("Unique Out") anomaly dataset a less challenging but more realistic benchmark. The disadvantage of this method is that we do end up filtering out objects that occur in multiple rooms that may very well be anomalous in a different scene type (e.g., chicken meat in the bathroom is an anomaly, but since "chicken" has been excluded from the set of possible anomalous objects, this will not be a datapoint in the test set). Details on both datasets can be found in Table \ref{table:a-dataset}.

Following this process, we were able to generate a Full version of 934,202 image and anomaly combinations for the Out-of-Scene dataset and 577,907 image and anomaly combinations for the Unique Out-of-Scene anomaly dataset. Fig. \ref{fig:anomaly-example} shows an example of an anomaly datapoint drawn from the Unique Out-of-Scene anomaly dataset. 

\begin{rem}
It is important to note that our anomaly datasets are not adding anomalies at the image level. Directly injecting objects into images without generating artifacts is both an extremely difficult task and unnecessary for the evaluation of our technique.
\end{rem}

When testing the models trained on the Filtered and Detector datasets, which did not include the full set of objects, any anomaly datapoints where an out-of-dataset object was used as the anomaly were removed from the dataset. Datapoints that simply contained an out-of-dataset object were left in the anomaly dataset, with said out-of-dataset objects ignored during testing. Additionally, any scenes with fewer than five objects remaining after filtering were also removed to ensure that there was a minimum amount difficulty in each anomaly datapoint.

\subsubsection{Data Splitting}

In many machine learning tasks, the training data and the evaluation objective are both tied to the same data modality, making it straightforward to create validation and test sets by randomly removing examples from the overall dataset. However, in our domain the training data was a knowledge graph and the evaluation data consisted of sets of objects containing anomalies. This setup prevented us from directly sampling validation and testing examples from the training dataset. Instead, we chose to separate our training, validation, and testing data at the Visual Genome image level. We split the corresponding Visual Genome images into training, validation, and testing sets using a 80/10/10 split for each scene type, and then combined the per scene sets. This guaranteed that at least a few of each scene type would be included in the validation and testing sets; since there are a few extremely common classes and a few uncommon classes, a purely random split would have guaranteed that some classes would be missing from the training and validation splits. Finally, the training knowledge graph was extracted from only the separated training images.

\subsection{Model Training and Comparisons}

For training models, our experiment procedures were as follows. First, we trained KGE models on our three object-scene knowledge graphs (Full, Filtered, and Detector), varying the type of KGE model used and searching over other training hyperparameters. See the Appendix for details on the search space, selected models, and link prediction metrics. 

The inference hyperparameters, namely the number of scene contexts considered, $m$, and the weighting between scene context and object context, $\alpha$, were tuned during the validation set evaluation. 

As mentioned above, there is work being done in solving the link prediction problem and finding graph-based anomalies \cite{Akoglu2014GraphBA}, and we take advantage of this framework; however, given that context-dependent anomaly detection is not a widely studied problem, there are no public datasets for this domain that can be used as benchmarks. Therefore, we did not include comparisons to other techniques in this paper.

\section{Experimental Results and Discussion\label{results}}

\begin{table}
\parbox{1.0\linewidth}{
    \centering
        \begin{tabular}{ |l|c|c|c|c|c|c|   }
             \cline{2-5}
             \multicolumn{1}{c|}{} & \multicolumn{2}{c|}{Top 1 Accuracy} & \multicolumn{2}{|c|}{Top 3 Accuracy} \\ %& \multicolumn{2}{|c|}{Performance on Data Subsets} \\
             \hline
             Model & Out  & Unique Out & Out & Unique Out   \\
             \hline
             Full     & 80.7  & 88.3  & 96.5  & 99.0   \\
             Filtered & 62.5  & 66.8  & 88.0  & 91.1  \\
             Detector & 97.0  & 99.3  & 99.7  & 99.9  \\
             \hline
            \end{tabular}
            \centering
            \caption{Anomaly prediction performance for each model and anomaly dataset}
            \hfill
            \vspace{-0.8cm}
            \label{table:acc}
    }
\end{table}
\begin{table}[t]
\parbox{1.0\linewidth}{
        \centering
        \begin{tabular}{ |l|c|cccc| }
             \hline
             True Scene & \vtop{\hbox{\strut Anomaly}\hbox{\strut Prediction}} & \multicolumn{4}{|c|}{Other Objects in Input}  \\
             \hline
             driveway& \textbf{broccoli} & window & line & road & car  \\
             \hline
             garden & milk & rosemary & jalapeno & \textbf{broccoli} & pile  \\
             \hline
            \end{tabular}
            \centering
            \caption{Example of correct identification of anomaly and non-anomaly.}
            \vspace{-0.9cm}
            \label{table:detectedexample}
        }
\end{table}
\begin{table*}        
\parbox{1.0\linewidth}{
        \centering
        \begin{tabular}{ |l|c|ccccc| }
             \hline
             True Scene & True Anomaly & \multicolumn{5}{|c|}{Higher Anomaly Score Objects}  \\
             \hline
             hallways & boards & mounted & elderly & cameras & escalator & terminal  \\
             \hline
             office & drink & mounted & sit  & thermostat  & hanging  & printer  \\
             \hline
             garden & straps & flown & polo & wind & streamers & pasture  \\
             \hline
            \end{tabular}
            \centering
            \caption{Examples of anomaly datapoints misidentified}
            \vspace{-0.6cm}
            \label{table:undetectedexample}
        }
\end{table*}

\subsection{Anomaly Detection}

 Table \ref{table:acc} provides an overview of our results. Our best performing model on the Full dataset correctly identified 80.7\% Out-of-Scene anomalies and 88.3\% of Unique Out-of-Scene anomalies. Relaxing the accuracy criteria to top 3 accuracy (the frequency with which the model ranked the true anomaly in the top three anomaly scores), we found that the detection rate jumped to 96.5\% and 99.0\%, respectively. Given that there were on average 20.55 objects in each anomaly scene, our model's performance was substantially greater than a random model.

\subsubsection{Qualitative Analysis of Anomaly Detection Performance}

When examining the decisions our models made, a few encouraging qualitative trends emerge. Objects that were correctly detected as anomalies in one scene were also correctly identified as non-anomalous in other inputs. Table \ref{table:detectedexample} shows an example of "broccoli" being correctly flagged in "driveway", and correctly not flagged in "garden."

\begin{figure}[t]
        \centering
        \includegraphics[width=1\columnwidth]{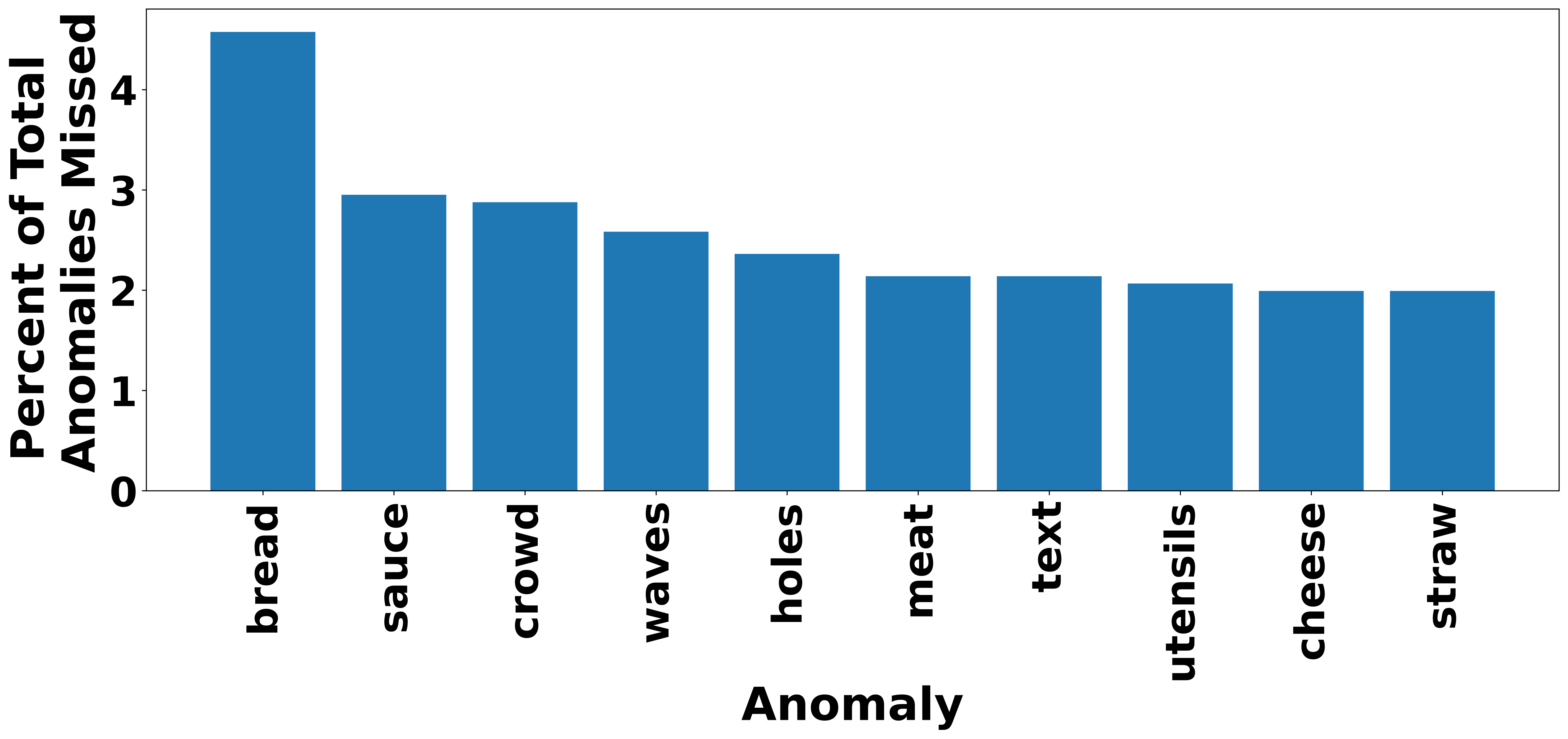}
        \caption{The 10 anomalies missed most frequently by the best performing Full model on the Unique Out-of-Scene anomaly dataset, as a percentage of the total anomalies that went undetected. Combined, they account for about 25\% of the total undetected anomalies. Note that most objects in the set are either food related or abstract.}
        \vspace{-0.1cm}
        \label{fig:canoms}
\end{figure}

Most of the errors the model made contained anomalies that fall into two categories: abstract objects like "text" or "holes," and food related objects like "meat" and "cheese." Fig. \ref{fig:canoms} includes the objects that most frequently escaped detection for top 5 accuracy when using the Unique Out-of-Scene dataset. Since both categories can be found in multiple scene types in the real world, their inclusion as anomalies likely stems from our limited data and automated anomaly generation procedure rather than a direct limitation of the model. 

Further support for this idea can be found in Table \ref{table:undetectedexample}, which shows examples of the other objects that were predicted to have higher anomaly scores than the anomalous object. These examples were randomly selected to avoid introducing bias, and similarly show that datapoints where the model was unable to flag the anomalous object contained other seemingly more anomalous objects (i.e., boards in hallways over escalators in hallways). While the model does make some genuine mistakes (i.e., rating straps as more similar to garden than pasture), overall this indicates that our model performance might increase on an anomaly dataset constructed under more human supervision. 

\subsubsection{Context Trade-off}
\begin{figure}[t]
    \centering
    \includegraphics[width=1\columnwidth]{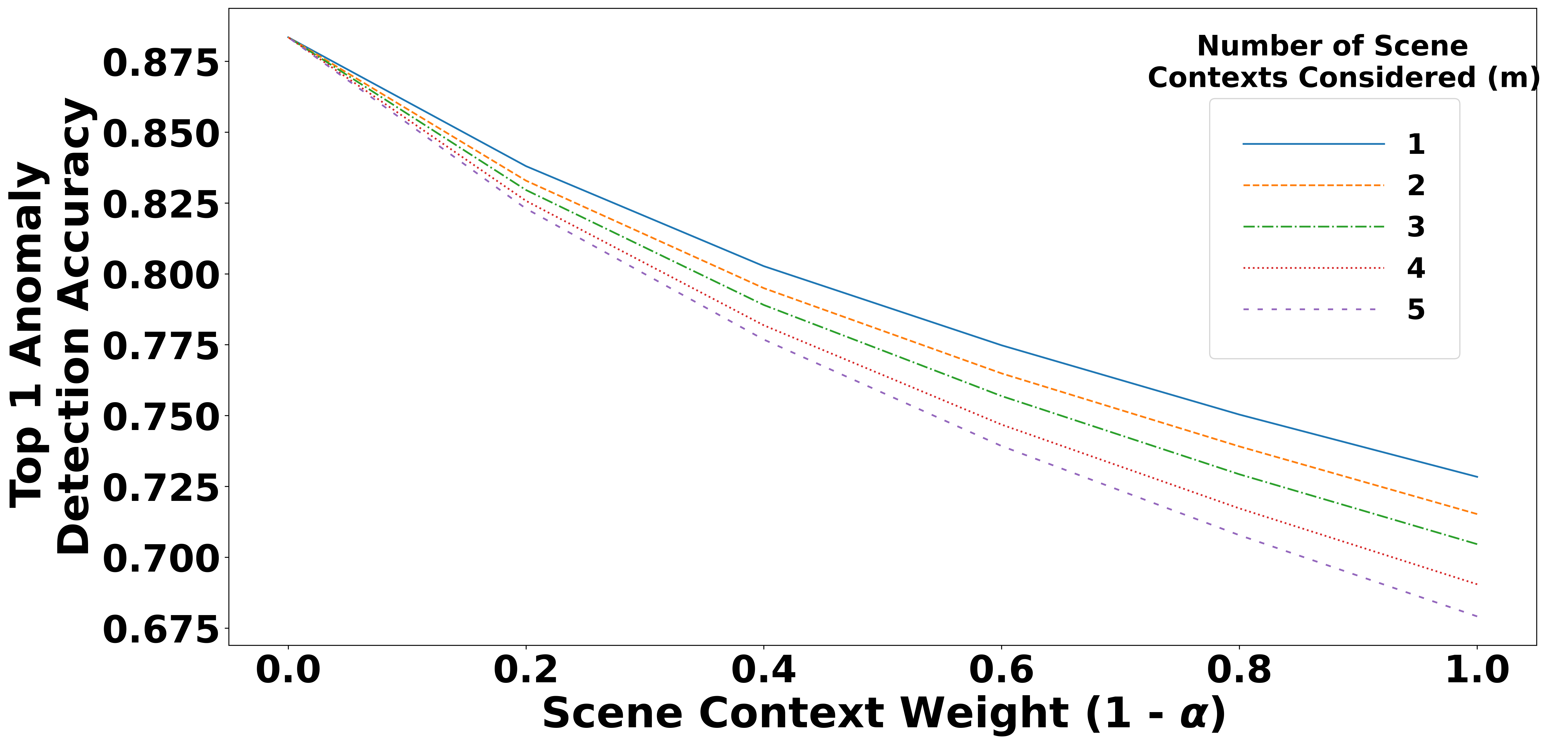}
    \caption{Decrease in performance as scene context is weighted more heavily relative to object context. Trend holds regardless of the number of scene contexts considered, but faster decreases in performance are seen with higher number of contexts. Highest performance is achieved when scene context is completely ignored.}
    \vspace{-0.6cm}
    \label{fig:ctradeoff}
\end{figure}

During the tuning of $\alpha$, we consistently found that our models achieved the best performance when we maximized contribution of object context, with performance decreasing steadily as the contribution of the the scene context increased. Fig. \ref{fig:ctradeoff} shows this trend for our best performing model, which held regardless of the number of scenes $m$ that were considered. To further explore this observation, we calculated the model's accuracy on scene prediction and found that it achieved a top 1 accuracy of 58.9\% and a top 5 accuracy of 80.4\%. While significantly lower than the accuracy achieved by state-of-the-art computer vision-based models, these results indicate that there is usable information coming from the scene context, and it is possible that future work will be able to better utilize this information.
\begin{table}[t]
\parbox{1.0\linewidth}{
    \centering
        \begin{tabular}{ |c|c|c|c|}
             \hline
             \multicolumn{2}{|c|}{Out} & \multicolumn{2}{|c|}{Unique Out} \\
             \hline
             \multicolumn{4}{|c|}{Model Top 1 Accuracy} \\
             \hline
             Filtered & Detector  & Filtered & Detector  \\
             84.0 & 95.5  & 92.2 & 98.8  \\
             \hline
             \hline
             \multicolumn{4}{|c|}{Model Top 3 Accuracy} \\
             \hline
             Filtered & Detector  & Filtered & Detector  \\
             96.9 & 99.7  & 99.4 & 99.9  \\
             \hline
            \end{tabular}
            \centering
            \caption{Anomaly prediction performance of Full model when tested on the Filtered and Detector Out-of-Scene and Unique anomaly datasets. Notice that the top 1 accuracy shown here on the Detector-based Out-of-Scene dataset is 95.5\%, compared to the Detector model's performance on the same dataset of 97\% (shown in Table \ref{table:acc})}
            \vspace{-1.0cm}
        \hfill
        \label{table:adjuperf}
    }
\end{table}
\hfill

\subsubsection{Performance When Trained on Detector Dataset}

The best model trained on the Detector dataset achieved a top 1 accuracy of 97.0\% and a top 3 accuracy of 99.7\% on the more challenging Out-of-Scene anomaly dataset. While these accuracy values are higher than those of the model trained on the Full dataset, the Detector-based anomaly dataset is a less difficult benchmark as there are only 9.31 objects per scene on average compared to the 20.55 objects per room in the Full dataset. However, since the Detector-based set is a subset of the Full set, models trained on the Full dataset can be tested against the Detector (and Filtered) anomaly dataset to provide a baseline for comparison. Table \ref{table:adjuperf} shows the accuracy values from this comparison, which found that there is only a 1.50\% difference in top 1 accuracy when looking at the performance of the models on the Detector Out-of-Scene dataset (Full model achieves 95.5\% and Detector model, shown in \ref{table:acc}, achieves 97.0\%), and no difference in top 3 accuracy (99.7\% for both Full and Detector models) between the two models. The very slight changes in performance between the Full model and the Detector model indicate that our methodology would be viable using state-of-the-art object detectors on raw image data, at least in terms of the raw number of types available.

\subsubsection{Performance When Trained on Filtered Dataset}

The models trained on the Filtered dataset performed significantly worse than the models trained on the Full, noisy dataset, with top 1 accuracy decreasing by 21.5\% and 25.4\% on the corresponding Out-of-Scene and Unique Out-of-Scene test sets. The most likely explanation of this result is that detrimental effects of the reduced amount of training data from filtering out noisy links outweighed any benefit gained from removing spurious correlations from the dataset. Interestingly, this decrease in performance was significantly larger than the decrease between the Full and Detector models. This is particularly notable since the Detector training dataset has a similar number of training links as compared to the Filtered dataset, which implies that there is be a set of filters that reduce the amount of training data required without overly lowering performance.

\section{Conclusion and Future Work}

In this paper, we demonstrate the use of a KGE-based
method for context-dependent anomaly detection that is
scalable, efficient, and interpretable. We successfully show that this method can identify anomalies in a single domain, household scenes, when given a relatively small amount of labeled image training data, and show that widely available object detector datasets provide enough classes to apply our method. Despite this success, there are several clear avenues in which our methods can be extended. Immediate work will include adding an object detector to tackle the challenge of building a knowledge graph directly from or performing inference on noisy image data and incorporating online updates. We are also exploring adding new relational links and implementing a graph neural network in the pipeline to leverage richer information about the graph structure and learn more complex relationships. This, paired with the object detector, will allow us to observe performance in point and group anomalies as well. Additionally, while our methodology for developing anomaly datasets was sufficient for this work, it could be improved with additional human oversight or better filtering techniques. Context-dependent anomaly detection is still in its infancy as a research direction, but is an interesting problem for continued exploration.

%\addtolength{\textheight}{-12cm}   % This command serves to balance the column lengths
                                  % on the last page of the document manually. It shortens
                                  % the textheight of the last page by a suitable amount.
                                  % This command does not take effect until the next page
                                  % so it should come on the page before the last. Make
                                  % sure that you do not shorten the textheight too much.

%%%%%%%%%%%%%%%%%%%%%%%%%%%%%%%%%%%%%%%%%%%%%%%%%%%%%%%%%%%%%%%%%%%%%%%%%%%%%%%%

%%%%%%%%%%%%%%%%%%%%%%%%%%%%%%%%%%%%%%%%%%%%%%%%%%%%%%%%%%%%%%%%%%%%%%%%%%%%%%%%

%%%%%%%%%%%%%%%%%%%%%%%%%%%%%%%%%%%%%%%%%%%%%%%%%%%%%%%%%%%%%%%%%%%%%%%%%%%%%%%%
\section*{Appendix}

\begin{table}[t!]
\parbox{1\linewidth}{
        \centering
        \begin{tabular}{ |l|c|   }
             \hline
               Parameter Type & Values \\ %& \multicolumn{2}{|c|}{Performance on Data Subsets} \\
             \hline
             Model & TransE TransR TransD ComplEx Analogy \\
             \hline
             Learning Rate & 5e-3 1e-3 5e-4 \\
             \hline
             Learning Rate Schedule & None Linear 1Cycle \\
             \hline
             Object Embedding Size & 25 50 75 100 500 700 800 1000 \\
             \hline
             Relation Embedding Size & 25 50 75 100 150 \\
             \hline
             Epochs & 10 50 100 200 300 400 500 1000 2000  \\
             \hline
            \end{tabular}
            \caption{Model and Hyperparameter Search Space Used for training link prediction models}
            \hfill
            \vspace{-0.3cm}
            \label{table:params}
    }
    \parbox{1\linewidth}{
        \centering
        \begin{tabular}{ |l|c|c|c| }
             \cline{2-4}
             \multicolumn{1}{c|}{} & \multicolumn{3}{c|}{Model} \\
             \hline
             Parameter Type & Full & Filtered & Detector \\ %& \multicolumn{2}{|c|}{Performance on Data Subsets} \\
             \hline
             Model & TransD & TransR & TransD \\
             \hline
             Learning Rate & 5e-3 & 1e-3 & 1e-4\\
             \hline
             Learning Rate Schedule & Linear & None & None  \\
             \hline
             Object Embedding Size & 75  & 300 & 400 \\
             \hline
             Relation Embedding Size & 75  & 150 & 100   \\
             \hline
             Epochs & 500 & 1000 & 1000  \\
             \hline
            \end{tabular}
            \caption{Model Type and Hyperparameters that resulted in best performing model for Each Dataset}
            \hfill
            \vspace{-0.7cm}
            \label{table:bestparams}
    }
\end{table}

\begin{table}
    \parbox{1\linewidth}{
        \centering
        \begin{tabular}{ |l|c|c|c| }
             \cline{2-4}
             \multicolumn{1}{c|}{} & \multicolumn{3}{c|}{Model} \\
             \hline
             Metric Type & Full & Filtered & Detector \\ %& \multicolumn{2}{|c|}{Performance on Data Subsets} \\
             \hline
             Filtered Hits @ 10 & 35.9  & 92.6  & 96.4 \\
             \hline
             Filtered Mean Rank & 154.8 & 4.69 & 3.33  \\
             \hline
             Filtered MRR & 0.144 & 0.429 & 0.439 \\
             \hline
            \end{tabular}
            \caption{Best Model Performance on Link Prediction Metrics for Each Dataset}
            \hfill
            \vspace{-1.1cm}
            \label{table:linkperf}
    }
\end{table}

Table \ref{table:params} shows the full space of hyperparameters and models which were tested in this work. Table \ref{table:bestparams} shows the best hyperparameter combinations for each training data set, and Table \ref{table:linkperf} shows the performance of each model on various common link prediction metrics. Note that the metrics are not comparable between datasets, as they depend on the size of the dataset.

\section*{Acknowledgments}

The authors would like to thank Dr. Zachary Serlin for his time and helpful feedback, and Drs. Rajmonda Caceres, Lori Layne, and Sung-Hyun Son for their support. 

%%%%%%%%%%%%%%%%%%%%%%%%%%%%%%%%%%%%%%%%%%%%%%%%%%%%%%%%%%%%%%%%%%%%%%%%%%%%%%%%

\bibliographystyle{IEEEtran}
\bibliography{ms}

\end{document}